\documentclass[conference]{IEEEtran}
\IEEEoverridecommandlockouts
\usepackage{cite}
\usepackage{amsmath,amssymb,amsfonts}
\usepackage{bm}
\usepackage{csquotes}
\usepackage{algorithmic}
\usepackage{graphicx}
\usepackage{textcomp}
\usepackage{subcaption}
\usepackage{threeparttable}
\usepackage{xcolor}
\usepackage{comment}
\usepackage{tablefootnote}
\def\BibTeX{{\rm B\kern-.05em{\sc i\kern-.025em b}\kern-.08em
   T\kern-.1667em\lower.7ex\hbox{E}\kern-.125emX}}

\begin{document}

\title{Generalization Ability of Feature-based Performance Prediction Models:\\ A Statistical Analysis across Benchmarks}
\author{\IEEEauthorblockN{Ana Nikolikj, Ana Kostovska, Gjorgjina Cenikj, Carola Doerr, and Tome Eftimov  }
\thanks{Ana Nikolikj (Email: ana.nikolikj@ijs.si), Gjorgjina Cenikj (Email: gjorgjina.cenikj@ijs.si) and Tome Eftimov (Email: tome.eftimov@ijs.si) are with Computer Systems Department, Jo\v{z}ef Stefan Institute, 1000 Ljubljana, Slovenia.}
\thanks{Ana Kostovska (Email:ana.kostovska@ijs.si) is with the Department of Knowledge Technologies, Jo\v{z}ef Stefan Institute, 1000 Ljubljana, Slovenia.}
\thanks{Ana Nikolikj. Gjorgjina Cenikj, and Ana Kostovska are also with the Jo\v{z}ef Stefan International Postgraduate School, 1000 Ljubljana, Slovenia.}
\thanks{Carola Doerr (Email: carola.doerr@lip6.fr) is with the Sorbonne Université, CNRS, LIP6, 1000 Paris, France.}
\thanks{The authors acknowledge the support of the Slovenian Research Agency through program grants P2-0098 and P2-0103, project grant J2-4460, and young researcher grant No. PR-12897 to AN, PR-12393 to GC, and PR-09773 to AK, and a bilateral project between Slovenia and France grant No. BI-FR/23-24-PROTEUS-001 (PR-12040).}
}

\maketitle

\begin{abstract}
This study examines the generalization ability of algorithm performance prediction models across various benchmark suites. Comparing the statistical similarity between the problem collections with the accuracy of performance prediction models that are based on exploratory landscape analysis features, we observe that there is a positive correlation between these two measures. Specifically, when the high-dimensional feature value distributions between training and testing suites lack statistical significance, the model tends to generalize well, in the sense that the testing errors are in the same range as the training errors. Two experiments validate these findings: one involving the standard benchmark suites, the BBOB and CEC collections, and another using five collections of affine combinations of BBOB problem instances.
\end{abstract}


\section{Introduction}
Automated algorithm configuration~\cite{prager2020per,BelkhirDSS17} and selection~\cite{jankovic2020landscape,kerschke2019automated} are gaining a lot of attention in evolutionary computation. In most cases, they are performed by using supervised Machine Learning (ML) predictive models that use the feature representation of a problem instance (i.e., its characteristics) as input data and predict the performance of a specific algorithm (instance) achieved on that problem instance. However, one of the main drawbacks presented in these learning tasks is the low generalization ability of the predictive models. That is, the models tend to fail to provide accurate predictions for problem instances whose feature representation is underrepresented or not presented in the training data. For example, \v{S}kvorc et al.~\cite{vskvorc2022transfer} showed that a random forest (RF) model trained on the BBOB suite of the COCO environment~\cite{hansen2021coco} yields poor results when tested on the artificially generated problem instances from~\cite{tian2020recommender} and vice-versa. Kostovska et al.~\cite{kostovska2022per} show that an automated algorithm selector which is based on performance prediction models trained on the BBOB benchmark suite, cannot generalize on problem instances that are part of the Nevergrad's YABBOB~\cite{bennet2021nevergrad} benchmark suite.

By using feature representation of problem instances, several studies~\cite{zhang2019similarity,vskvorc2020understanding,munoz2020generating,eftimov2020linear,selector,long2022bbob} perform complementary analyses of different benchmark suites in the feature space. However, all the analyses are descriptive, trying to understand the similarities and differences between the problem instances across different benchmark suites without quantifying the similarities on a benchmark suite level. 
Nikolikj et al.~\cite{10.1145/3583133.3590617} have explored how well a performance predictive model can adapt based on the benchmark suite coverage. Empirical meta-features for each suite have been created by clustering instances across all suites and examining their similarities. The findings indicated that when two benchmark suites share similar empirical coverage, an ML model trained on one can perform well on the other.

\textbf{Our contribution:} In this study, we investigate the generalization ability of a performance prediction model through a statistical measure assessing the similarity of coverage among benchmark suites. Unlike the previous published empirical approach, which involved condensing high-dimensional benchmark data into a lower-dimensional space using clustering to define meta-representations for each suite, we directly utilize the raw benchmark suite data from the high-dimensional space -- representing all instances with meta-features. Employing a statistical test allows us to compare suite coverage distributions in their original high-dimensional space without losing information through conversion to a lower-dimensional space, as done previously. To assess patterns between the feature landscape and performance realms, we trained a predictive model for a specific optimization algorithm on one benchmark suite and evaluated it on another. The results imply that statistical insights from the feature landscape can anticipate how well a model extends to various suites. When the high-dimensional feature landscape distributions of training and testing suites are not statistically significant, the model archives good performance on the testing suite preserving an error within the training error range. These conclusions arise from two experiments: one encompassing typical benchmark suites for algorithm evaluation and another that employed sampling to generate five new artificial benchmark suites from problem instances.

\textbf{Outline:} Section~\ref{sec:related_work} presents an overview of complementary analyses performed across different benchmark suites. Section~\ref{sec:gen_workflow} introduces the workflow used to estimate and evaluate the generalization ability of a predictive model across different benchmark suites. The experimental design is explained in Section~\ref{sec:experiments}, followed by a discussion of key results in Section~\ref{sec:results}. Section~\ref{sec:conclusion} concludes the paper.

\textbf{Data and code availability:} The data and the code involved in this study are available at~\cite{GitRepository}.

\section{Related work}
\label{sec:related_work}
Most of the studies performed in the direction of performance prediction of single-objective black-box optimization algorithms rely on Exploratory Landscape Analysis (ELA)~\cite{mersmann2011exploratory} to calculate features that describe the properties of the problem instances. ELA is a set of mathematical and statistical techniques that use a sample of candidate solutions from the problem instance decision space, generated using a certain sampling technique. They can be calculated using the R programming language package called ``flacco"~\cite{flacco}. A recent version of the package has also been published in Python~\cite{prager2023pflacco}. These features have been used in several studies for complementary analysis between the benchmark suites.

Zhang and Halgamuge analyze the similarity of continuous problem instances by representing them with algorithm performance~\cite{zhang2019similarity}. The results indicate that the problem instances from different problem classes exhibit similarities in performance and that low-dimensional instances could also share performance similarities with their high-dimensional counterparts. \v{S}kvorc et al.~\cite{vskvorc2020understanding} analyze the complementary of BBOB and CEC benchmark suites by representing the problem instances using ELA features and further visualizing them with the t-distributed stochastic neighbor embedding (tSNE) method~\cite{gisbrecht2015parametric} in lower dimensions. The results show that the benchmark suites have different distributions over the landscape feature space. Mu{\~n}oz and Smith Miles~\cite{munoz2020generating} use genetic programming to generate new problem instances with controllable characteristics to increase the coverage of the problem landscape. The results demonstrate that the newly generated problem instances are more challenging for the algorithms to solve than the well-known benchmark suites. Eftimov et al.~\cite{eftimov2020linear} perform a correlation analysis of the projection of the ELA feature representations into the subspace obtained by a singular value decomposition, between the BBOB problem instances and the HappyCat and HGbat problem instances. Cenikj et al.~\cite{selector} present an approach, SELECTOR, for selecting diverse problem instances based on their ELA feature representation. They evaluate different sampling heuristics, one based on clustering and two based on graph embeddings. The results show that regardless of the choice of sampling heuristic, the approach leads to a reproducible statistical comparison of algorithm performance. Long et al.~\cite{long2022bbob} have provided a detailed analysis of landscape properties and algorithm performance across BBOB problem instances.

Previous studies have explored the complementarity of benchmark suites empirically, without directly linking it to predictive model generalization. A recent study found that when benchmark suites share similar empirical coverage~\cite{10.1145/3583133.3590617}, training a model on one suite yields good generalization on another. However, this depends on the meta-representation used. Our study addresses this by examining a statistical measure based on raw problem landscape data.

\section{Statistical measure for accessing similarity of benchmark suites}
\label{sec:gen_workflow}
Consider a scenario involving $m$ benchmark suites, each comprising varying numbers of problem instances. From this pool, one of the $m$ suites is selected to serve as the training set for the supervised ML predictive model ($\mathcal{M}$), while the remaining $m-1$ suites are utilized for testing purposes. To evaluate the model $\mathcal{M}$'s generalization ability across diverse benchmark suites used for testing, we outline the following workflow:

\noindent1) Establishing a unified meta-representation at the individual problem instance level involves characterizing instances across all benchmark suites using a shared set of $n$ meta-features, describing their landscape properties. This approach ensures that all selected problem instances are mapped into the same $n$-dimensional vector space. With this representation, each benchmark suite can be represented as a matrix $BS_{k\times n}$, where $k$ is the number of instances that are part of a benchmark suite (which can be different for different benchmark suites).

\noindent2) Once each benchmark suite is represented by its matrix, we can use a statistical test to compare the high-dimensional coverage distributions between two of them (one used for training and the other for testing). To this end, a statistical test for comparing high-dimensional distributions should be utilized. A category of consistent, distribution-free tests applicable to high-dimensional spaces relies on nearest neighbors using the Euclidean distance metric~\cite{henze1988multivariate,schilling1986multivariate}. Szekely and Rizzo introduced the multivariate $\mathcal{E}$ test, demonstrating its universal consistency against all alternatives (not necessarily continuous) possessing finite second moments. Notably, the computational complexity of this test remains independent of dimensionality or sample size, making it a formidable contender among nearest-neighbor tests. Findings highlighted in~\cite{szekely2004testing} suggest that the multivariate $\mathcal{E}$ test stands out as one of the most robust tests available for analyzing high-dimensional data, which makes it a good choice for our analysis. 
    
    Let us assume that two benchmark suites are involved, ${P}_{k_1\times n}$ and ${Q}_{k_2\times n}$, where $p_{1}, p_{2}, \dots, p_{k_1}$ and $q_{1}, q_{2}, \dots, q_{k_2}$ are problem instances represented by $n$ landscape meta-features (i.e., vectors in $\mathbb{R}^n$) that belong to the two benchmark suites respectively. The multivariate $\mathcal{E}$ test statistic between them is defined as: 
 \begin{align}
\mathcal{E}_{k_1,k_2}&=\frac{k_1k_2}{k_1+k_2}\Big(\frac{2}{k_1k_2}\sum_{i=1}^{k_1}\sum_{m=1}^{k_2}\vert\vert p_i - q_m\vert\vert\nonumber\\
&-\frac{1}{k_1^2}\sum_{i=1}^{w_1}\sum_{j=1}^{w_1}\vert\vert p_i - p_j\vert\vert\nonumber\\
&-\frac{1}{k_2^2}\sum_{l=1}^{k_2}\sum_{m=1}^{k_2}\vert\vert q_l - q_m\vert\vert \Big ).
\end{align}
The initial double sum in the equation above indicates the distance between problem instances from the benchmark suites, while the subsequent double sums delineate the internal distances within each benchmark suite ($P$ and $Q$). The test statistic follows a degenerate two-sample V-statistic~\cite{leucht2013dependent} -- readers interested in the mathematical details of the test are invited to confer~\cite{szekely2004testing}.

 \noindent3) The outcome of this comparison yields a p-value, serving as an indicator to ascertain whether a difference exists or not in the coverage distribution between the two benchmark suites. If there is no statistical difference, there is a high likelihood that a performance predictive model trained on one of the benchmark suites can be also utilized and generalize the results on the other benchmark suite and vice-versa.

\section{Experimental design}
\label{sec:experiments}
We begin this section by detailing the benchmark suites chosen for conducting two experiments. Additionally, we include details on the performance and problem landscape data and we provide details on training the performance prediction models. 

\textbf{Benchmark suites}:
As mentioned before, we conduct two experiments where we perform statistical analysis of the generalization ability of the performance prediction models across various benchmark suits. 

\textbf{First experiment:} 
The first experiment statistically assesses the generalization ability of predictive models across four widely-used benchmark suites: BBOB (COCO)\cite{hansen2021coco}, CEC 2013\cite{cec2013}, CEC 2014~\cite{cec2014}, and CEC 2015~\cite{cec2015}. BBOB comprises 24 problem classes, from which we use five instances each. The CEC 2013, CEC 2014, and CEC 2015 benchmark suites contain 28, 30, and 15 problems, respectively. Finally, for CEC 2013, we ended up with 27 problems. This is because three problems (specifically, the 3rd, 7th, and 20th problems) were excluded due to missing data arising from the landscape feature calculation process, as detailed later in this section. The experiment considers problems with $D = 10$ numerical decision variables.

\textbf{Second experiment:} 
The second set of experiments uses statistical analyses to assess how predictive models generalize on artificially generated benchmark suites. Demonstrating the impact of a more strategic training data selection, we generate these suites by using instances created as affine recombinations of pairs of BBOB problem instances, as introduced in~\cite{dietrich2022increasing}. The experiment focuses on a fixed problem dimensionality of $D = 5$. This choice allows us to re-use available performance data.

To ensure a representative and diverse set of problem instances from those generated through affine recombinations, we apply the SELECTOR methodology~\cite{selector}. This involves converting benchmark problem instances into a graph format, where nodes represent individual problems and an edge is created if the cosine similarity between their meta-representations is 0.9 or higher. The Maximal Independent Set (MIS) algorithm~\cite{ghaffari2016improved} is used to select instances, ensuring diversity by making sure selected instances have a pairwise cosine similarity less than 0.9. Since the MIS algorithm is stochastic, we repeat the process five times, resulting in five benchmark suites: BS1, BS2, BS3, BS4, and BS5. They contain 56, 57, 56, 55, and 53 problem instances respectively, with minimal overlap between instances across the benchmarks. The sole instance of overlap occurs between the first and second benchmark suites and between the third and fifth benchmark suites, each involving a single problem instance.

\textbf{Performance data:} For the first experiment, we analyze performance data from a portfolio of three algorithms, the Covariance Matrix Adaption Evolutionary Strategy (CMA-ES)~\cite{hansen2001self_adaptation_es}, the Real Space Particle Swarm Optimization (PSO)~\cite{kennedy1995pso}, and of Differential Evolution (DE)~\cite{storn1997differential}, respectively. Their implementations are taken from the Nevergrad library~\cite{nevergrad}, with each algorithm being configured to its default hyper-parameter setting. We fix both the budget and the target. The computational budget for executing the algorithms is limited to 100,000 function evaluations. We also set a target precision threshold at $10^{-8}$. The algorithm terminates upon either exhausting its allocated budget or when achieving the target precision, defined as the absolute difference $f(x^{\text{best}})-f^*$ between the quality of the best-found solution $x^{\text{best}}$ and that of a global optimum $f^*:=\inf_x f(x)$. The experiments are run using the IOHexperimenter~\cite{iohexperimenter} environment, for convenience of accessing the BBOB functions and for logging the search trajectories in a standardized way. Due to the stochastic nature of the algorithms, 30 independent runs of each algorithm on each problem instance have been performed. Finally, the median target precision across 30 repetitions has been calculated. 

In the second experiment, the same algorithms are used, this time evaluated on the affine functions and with a budget of 10,000 function evaluations. Similar to the first experiment, 25 independent runs have been performed of each algorithm on each problem instance and we calculate the median target precision across 25 repetitions.

\textbf{Problem landscape data:} The landscape characteristics of the problems are represented using publicly available ELA features. Specifically, for the first experiment, we utilize the 64 ELA features available from~\cite{lang2021exploratory}. These feature values were computed using the Improved Latin Hypercube Sampling (iLHS) method~\cite{beachkofski2002improved_lhs}, with a sample size of $800D=8,000$ and repeated 30 times. The median value of each feature across 30 repetitions was calculated and used. The choice of a larger sample size was deliberate to minimize the randomness inherent in the feature extraction process. For the second experiment, we utilized the 14 ELA features available from~\cite{dietrich2022increasing}. The calculation of these feature values was carried out employing Sobol' sampling, with a sample size of $250 \times D=1,250$, and repeated 30 times. Same as the first experiment, the median value of each feature across 30 repetitions was calculated and used. 

\textbf{Comparing benchmark suite distribution:} The statistical comparisons have been performed using the R programming language. To compare the distributions of high-dimensional data the \textit{multivariate $\mathcal{E}$ test} is used, which is a part of the \enquote{energy} package~\cite{energy}.
 
\textbf{Predictive models:} For each benchmark suite, we train a Random Forest (RF) regression model to predict the algorithm's performance, measured by the target precision achieved with the 100,000 (first experiment) and 10,000 function evaluations (second experiment), respectively. Instead of predicting the median target precision for each problem instance in the original space, we train the models in log space. This is the value we are predicting with the ML models. We use the default implementation of the RF regressor from the \textit{scikit-learn} package in Python. The performance of each of the trained models is evaluated on the other benchmark suites (that have not been used for training the model) and we report the median absolute error (MDAE) across all problem instances in the test benchmark suite. We analyze the obtained results to examine whether the patterns identified in the coverage matrix are similarly reflected in the performance of the automated algorithm performance prediction model.
For this experiment, the feature values are scaled by subtracting the mean and scaling to unit variance, using the \textit{scikit-learn} package. The parameters used for the scaling are learned using the training suite and then applied to the test suite.

\section{Results and discussion}
\label{sec:results}
\subsection{First experiment}
Table~\ref{tab:stat_bbob_cec} displays the p-values obtained from the statistical comparison of feature-space distributions across various pairs of benchmark suites. Here, we only consider the feature space of the problem, disregarding the performance of the algorithms. 

\begin{table}[t]
\centering
\scriptsize 
{
\caption{Comparative statistical analysis of high-dimensional feature-space distributions between paired benchmark suites, scaling based on the collection listed in the row. Presenting p-values; values $\le .005$ are marked with an *. }
\label{tab:stat_bbob_cec}
\begin{threeparttable}
\begin{tabular}{|r|cccc|}
  \hline
 & BBOB &  CEC2013 & CEC2014 & CEC2015  \\ 
  \hline
     BBOB  & / &0.005$^*$ & 0.005$^*$& 0.005$^*$  \\ 
  CEC2013  & 0.035$^*$ & / & 0.105 & 0.005$^*$  \\ 
  CEC2014  & 0.005$^*$& 0.245& / & 0.690  \\ 
  CEC2015  & 0.005$^*$& 0.205 & 0.490 & /  \\ 
   \hline
\end{tabular}
\end{threeparttable}
}
\end{table}

Note also that we consider here pairwise comparisons, not multiple ones. 
The comparison matrix is not symmetric, caused by the scaling procedure explained above (which depends on the training set, i.e., here the set in the row). 

From this table, we observe that the feature-space distribution of the \textbf{BBOB} benchmark suite exhibits statistical significance when (individually) compared to CEC2013, CEC2014, and CEC2015, respectively, suggesting that a performance predictive model trained on BBOB data will likely yield higher errors than the training error when utilized for predictions on CEC2013, CEC2014, and CEC2015. For the CEC2013 suite, no statistical significance is observed when compared to CEC2014, however, statistical significance has been noted when compared to the BBOB and the CEC2015 suites. These findings suggest that a model trained using CEC2013 will demonstrate good predictive performance when applied to CEC2014. However, the prediction errors are likely to increase when this model is used for predictions on BBOB and CEC2015 benchmark suites. For \textbf{CEC2014} and \textbf{CEC2015}, please confer the table.

\begin{figure*}[t]
\centering
\begin{subfigure}[b]{0.32\textwidth}
   \includegraphics[width=\linewidth]{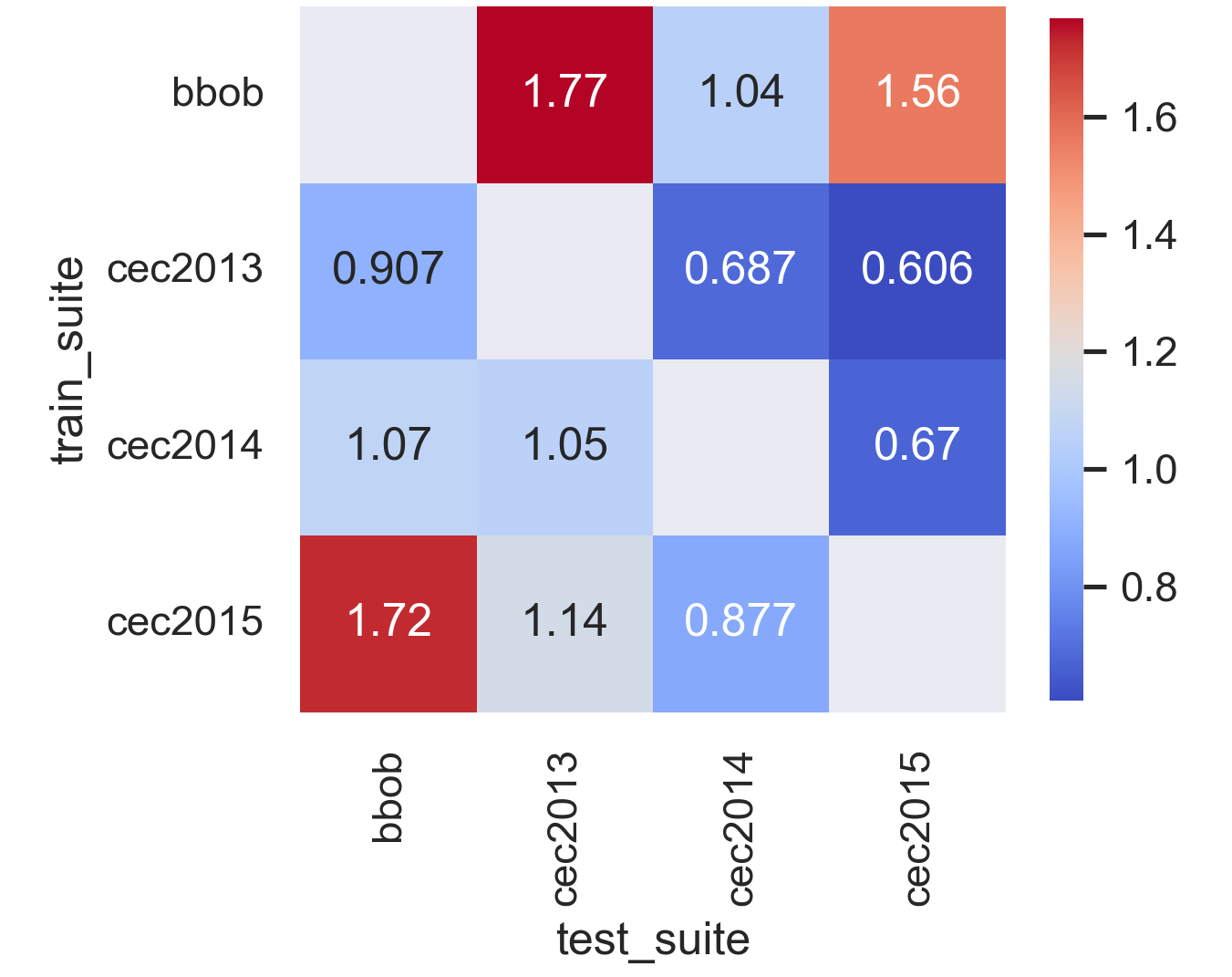}
   \caption{CMA}
   \label{fig:CMA} 
\end{subfigure}
\begin{subfigure}[b]{0.32\textwidth}
   \includegraphics[width=\linewidth]{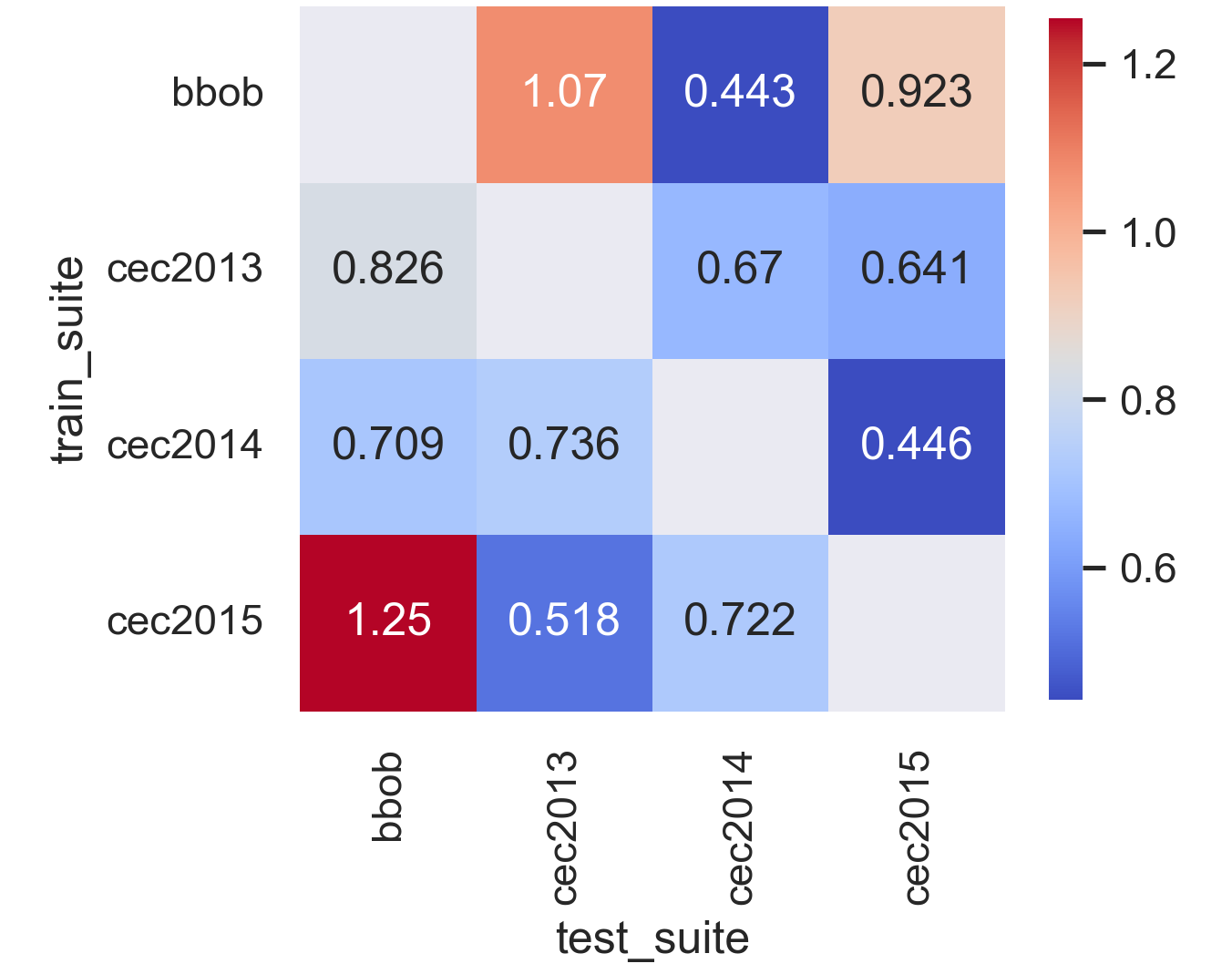}
   \caption{DE}
   \label{fig:DE}
\end{subfigure}
\begin{subfigure}[b]{0.32\textwidth}
   \includegraphics[width=\linewidth]{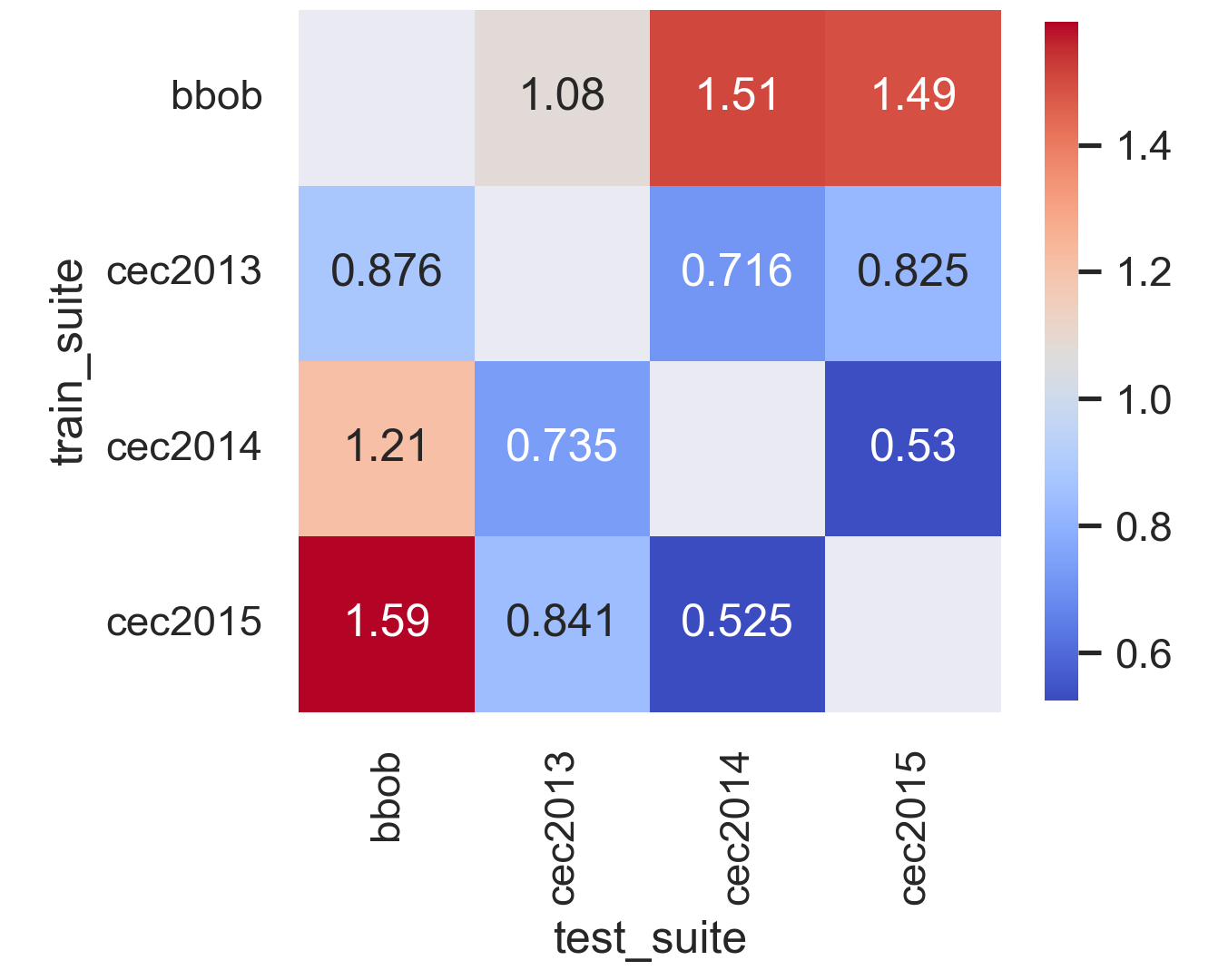}
   \caption{PSO}
   \label{fig:PSO}
\end{subfigure}
\caption{Heatmap showing the MDAE of an RF model when predicting the performance of a) CMA, b) DE, and c) PSO, on BBOB, CEC2013, CEC2014, CEC2015, and CEC2017. Rows indicate the training benchmark suite and columns indicate the benchmark suite of the model
was evaluated on.}
\vspace{-5mm}
\label{fig:pred_errors}
\end{figure*}

\begin{figure*}[t]
\centering
\begin{subfigure}[b]{0.23\textwidth}
   \includegraphics[width=\linewidth]{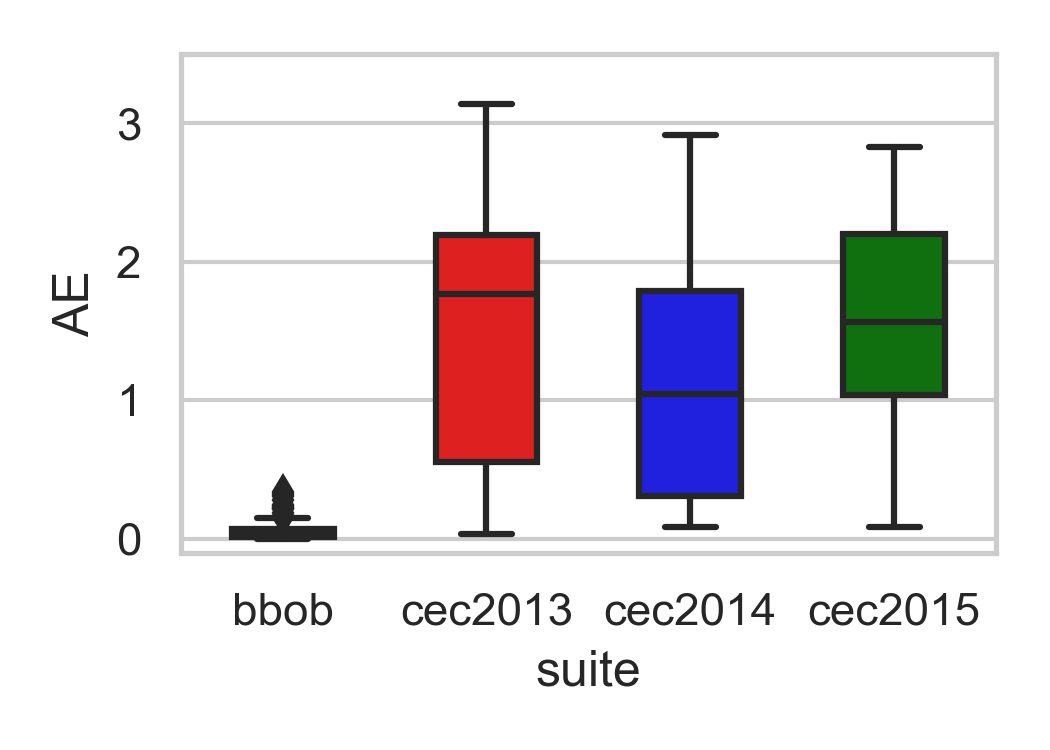}
   \caption{bbob}
   \label{fig:CMA} 
\end{subfigure}
\begin{subfigure}[b]{0.23\textwidth}
   \includegraphics[width=\linewidth]{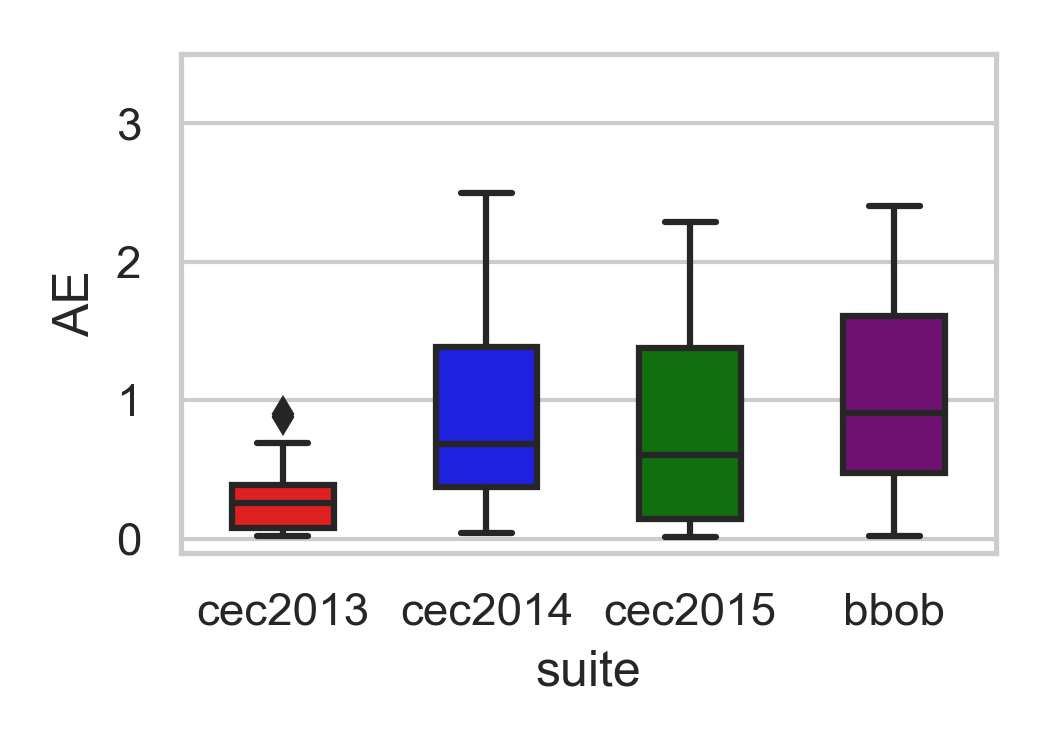}
   \caption{cec2013}
   \label{fig:DE}
\end{subfigure}
\begin{subfigure}[b]{0.23\textwidth}
   \includegraphics[width=\linewidth]{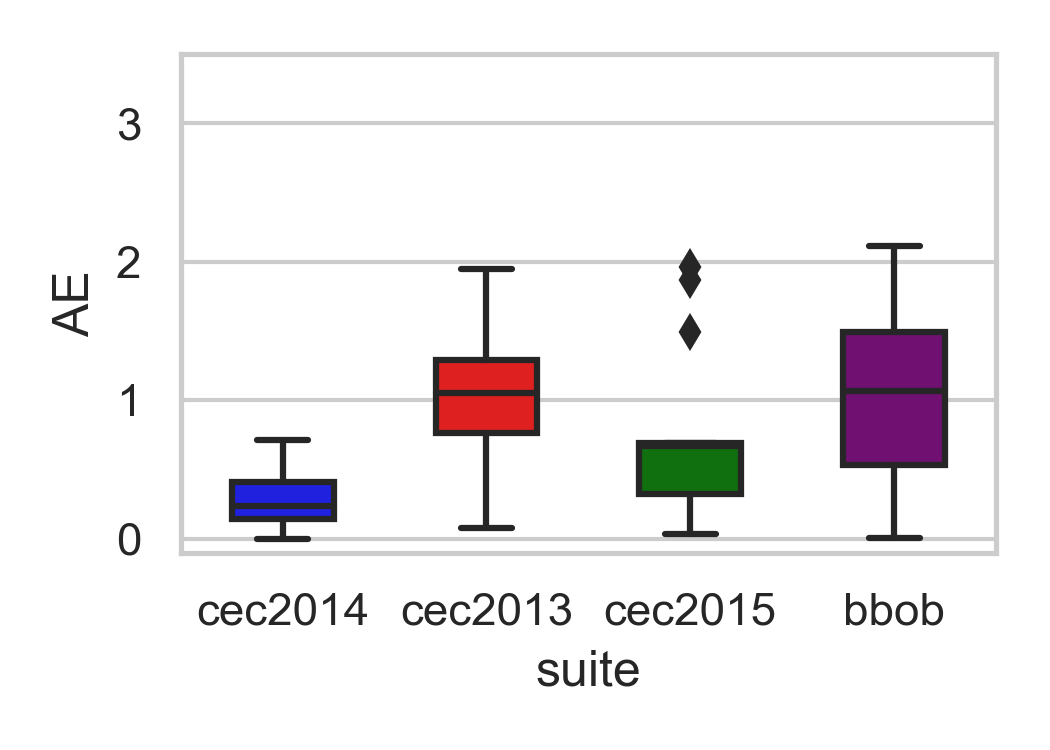}
   \caption{cec2014}
   \label{fig:PSO}
\end{subfigure}
\begin{subfigure}[b]{0.23\textwidth}
   \includegraphics[width=\linewidth]{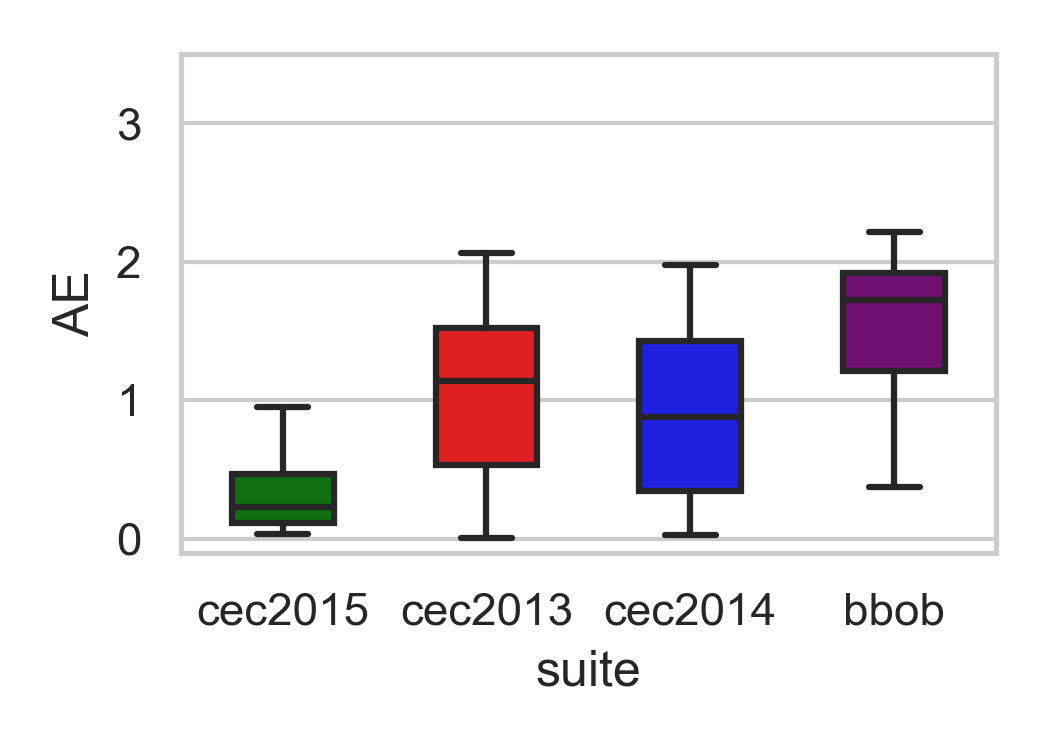}
   \caption{cec2015}
   \label{fig:PSO}
\end{subfigure}
\newline
\begin{subfigure}[b]{0.23\textwidth}
   \includegraphics[width=\linewidth]{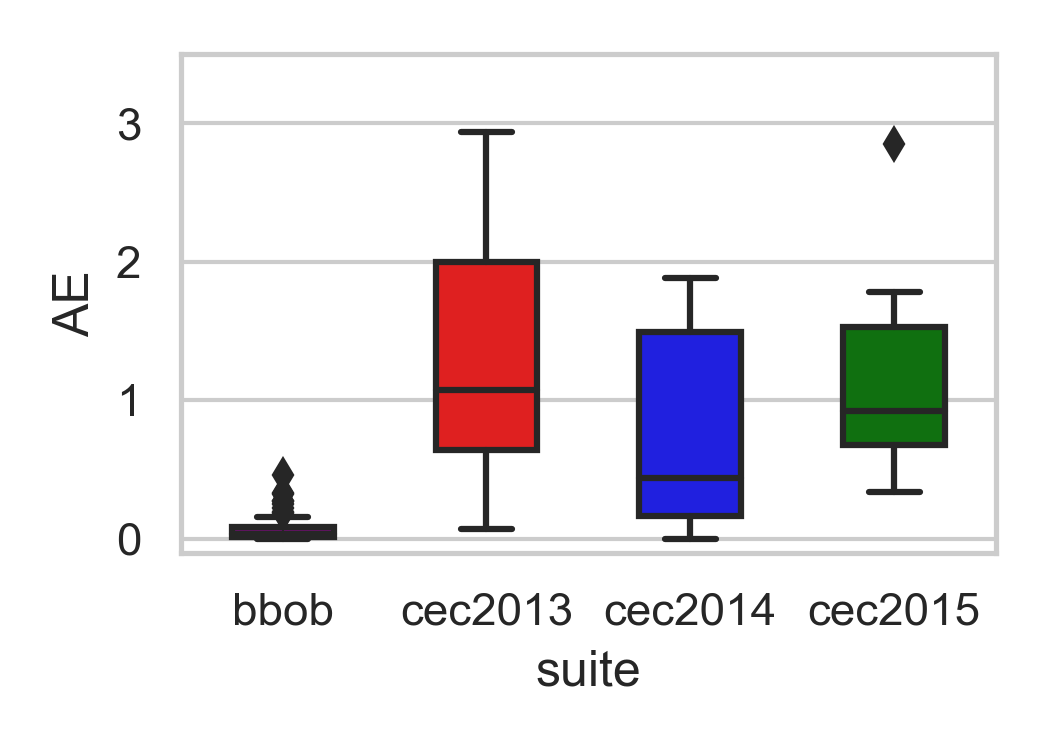}
   \caption{bbob}
   \label{fig:CMA} 
\end{subfigure}
\begin{subfigure}[b]{0.23\textwidth}
   \includegraphics[width=\linewidth]{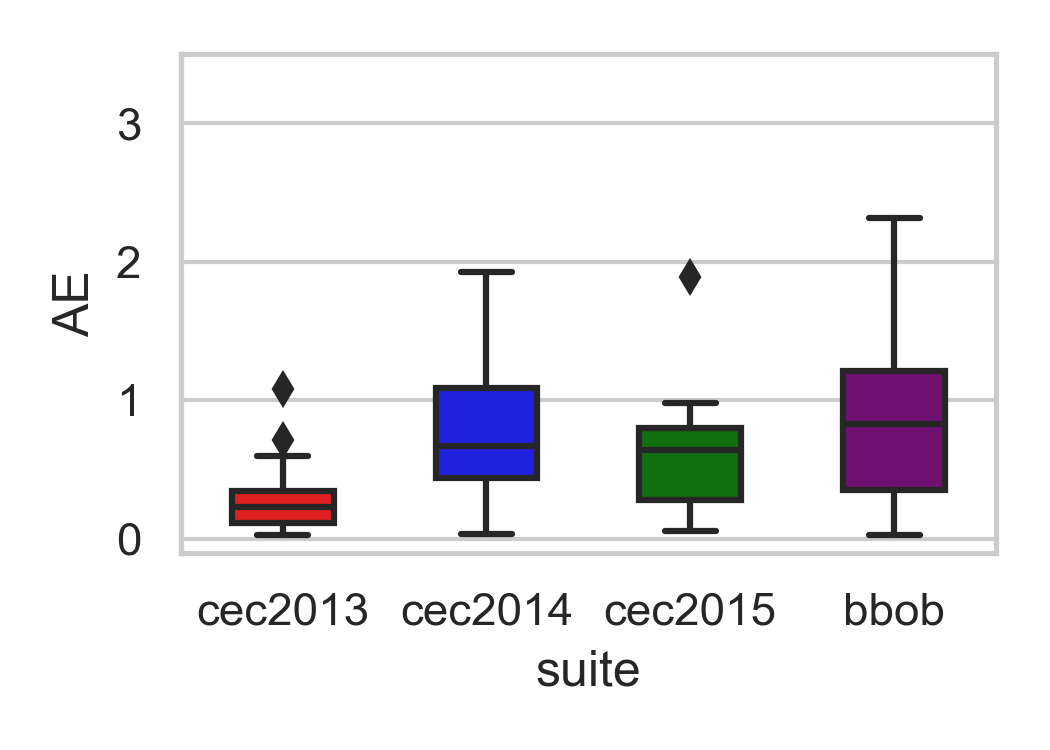}
   \caption{cec2013}
   \label{fig:DE}
\end{subfigure}
\begin{subfigure}[b]{0.23\textwidth}
   \includegraphics[width=\linewidth]{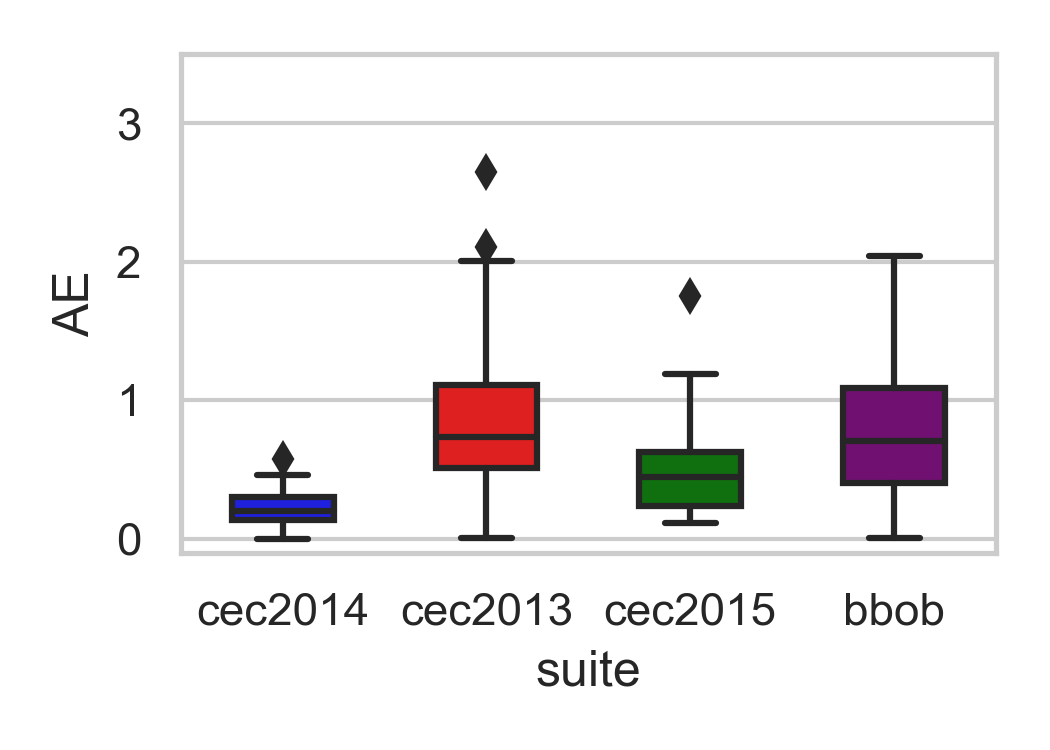}
   \caption{cec2014}
   \label{fig:PSO}
\end{subfigure}
\begin{subfigure}[b]{0.23\textwidth}
   \includegraphics[width=\linewidth]{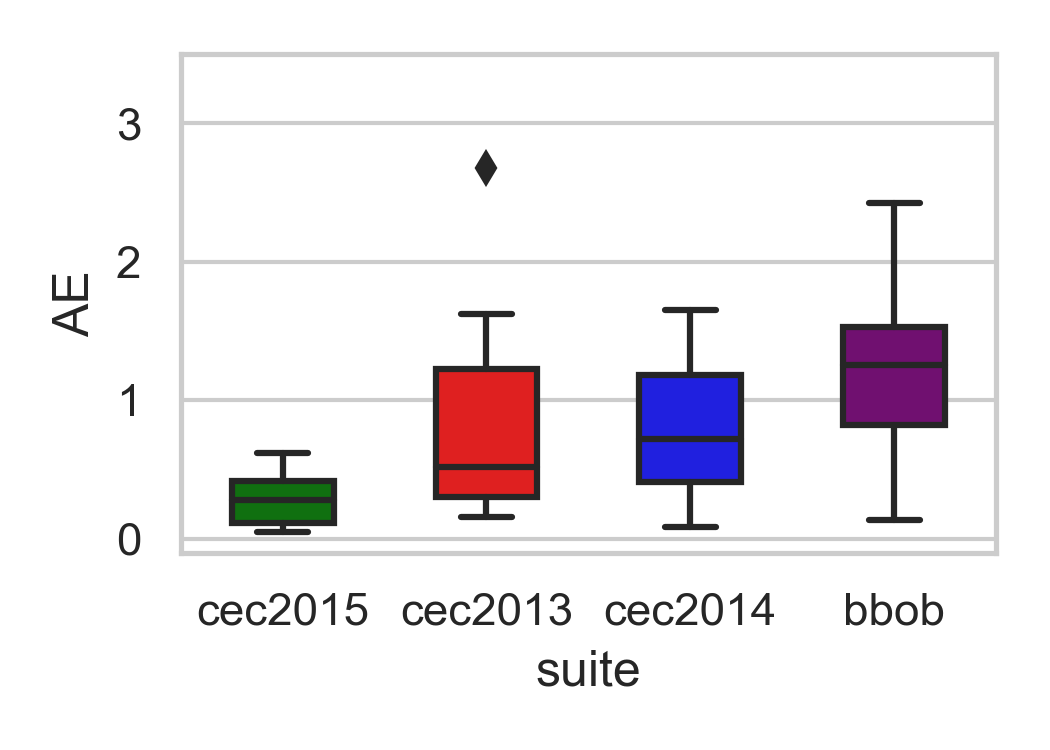}
   \caption{cec2015}
   \label{fig:PSO}
\end{subfigure}
\newline
\begin{subfigure}[b]{0.23\textwidth}
   \includegraphics[width=\linewidth]{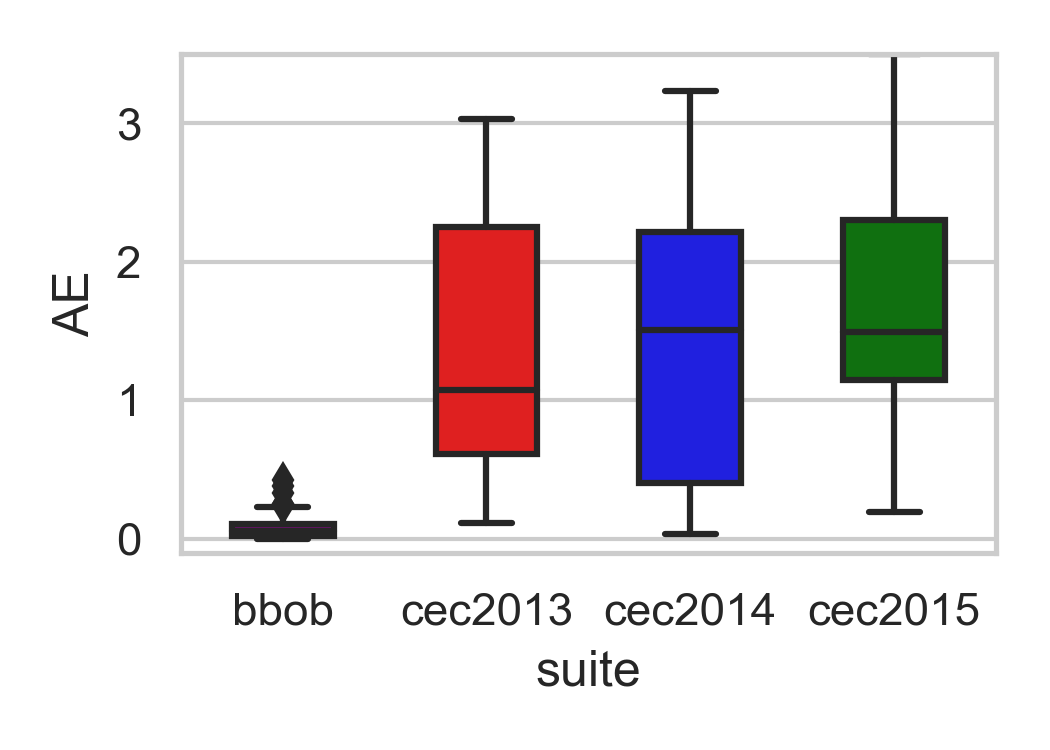}
   \caption{bbob}
   \label{fig:CMA} 
\end{subfigure}
\begin{subfigure}[b]{0.23\textwidth}
   \includegraphics[width=\linewidth]{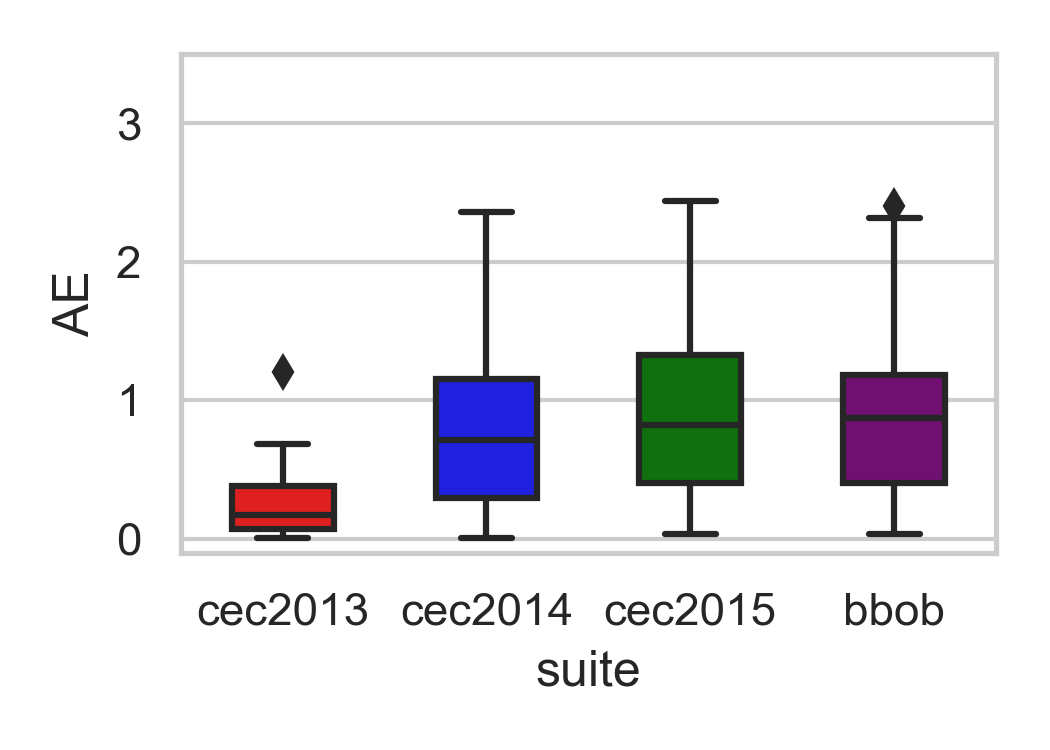}
   \caption{cec2013}
   \label{fig:DE}
\end{subfigure}
\begin{subfigure}[b]{0.23\textwidth}
   \includegraphics[width=\linewidth]{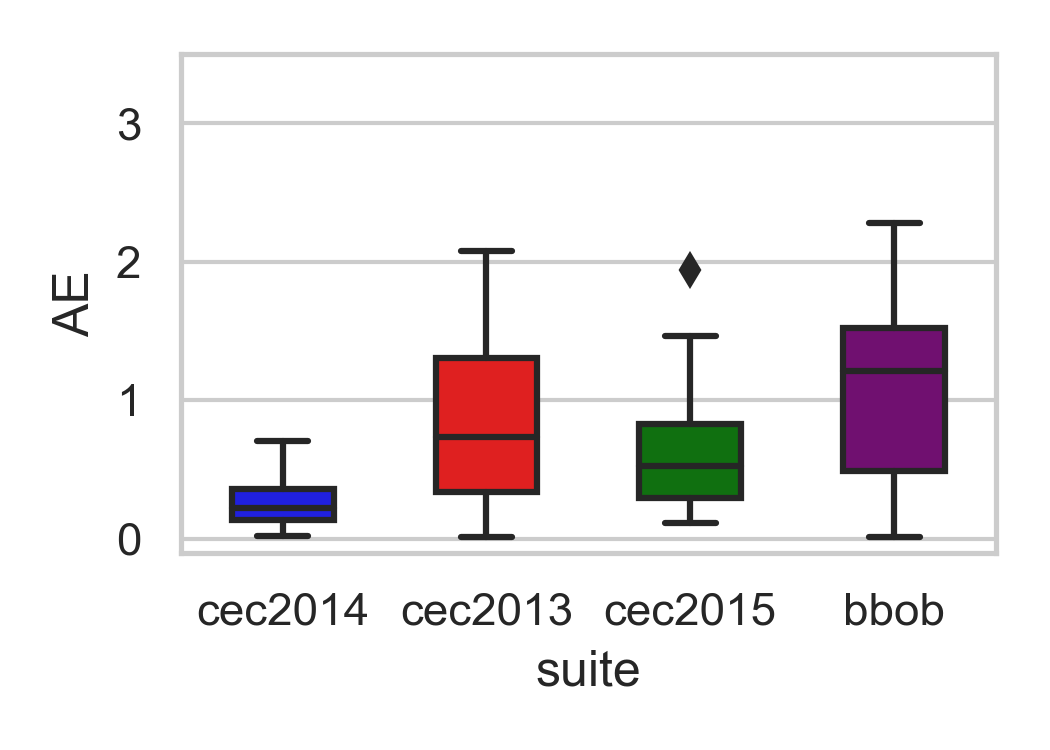}
   \caption{cec2014}
   \label{fig:PSO}
\end{subfigure}
\begin{subfigure}[b]{0.23\textwidth}
   \includegraphics[width=\linewidth]{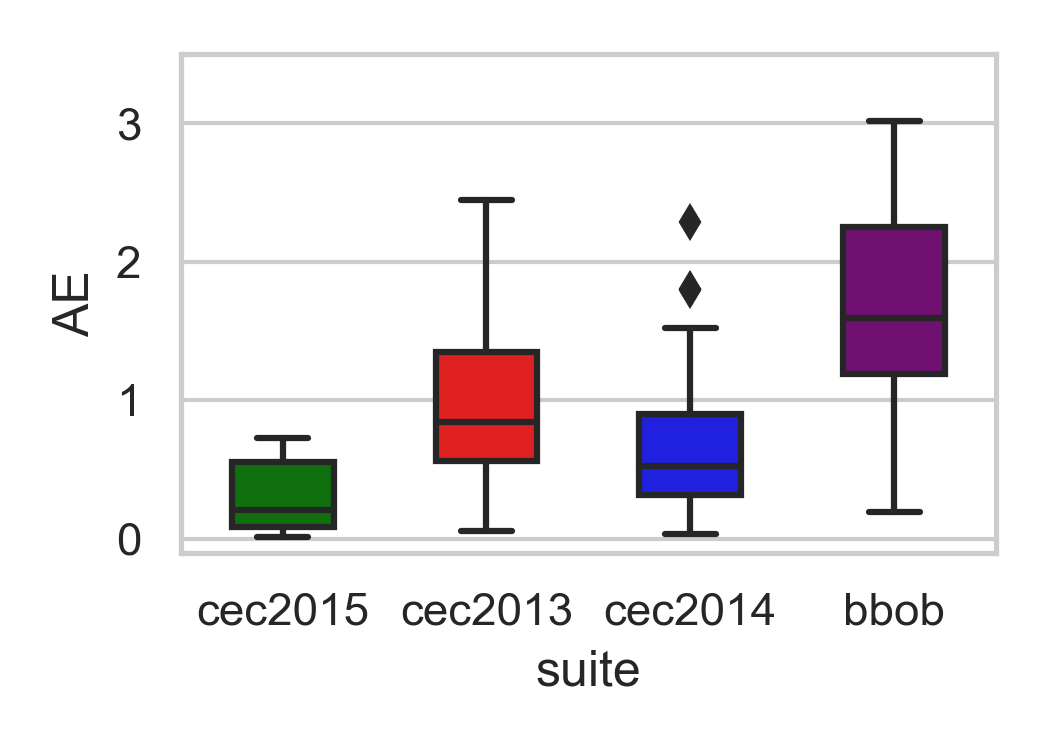}
   \caption{cec2015}
   \label{fig:PSO}
\end{subfigure}
\caption{Box-plots showing the AE (Absolute error) of an RF model when predicting the performance of a-d) CMA, e-h) DE, and i-l) PSO. Subplot titles name the training benchmark suite, with one box plot showing train AEs and others depicting corresponding test AEs.}
\vspace{-5mm}
\label{fig:pred_errors_boxplots}
\end{figure*}

To determine if the statistical patterns observed in the feature landscape space are consistent in the performance space, Fig~\ref{fig:pred_errors} showcases the MDAE (calculated as the median of the absolute differences between the predicted values and the ground truth value (i.e., log from the median target precision)) of the RF predictive model. This model is trained using one benchmark suite indicated in the rows of the heatmap and tested on the remaining three benchmark suites (columns of the heatmap). This analysis is conducted independently for three algorithms (CMA-ES, DE, and PSO). Additionally, Table~\ref{tab:train_errors_exp_1} displays the training errors of the RF models for each benchmark suite individually, so we can further analyze if the testing errors are in the same ranges as the training errors.

\begin{table}[t]
\centering
\caption{
The MDAE during training of the RF model within every benchmark suite for the algorithms CMA-ES, PSO, and DE.}
\label{tab:train_errors_exp_1}
\begin{tabular}{|r|cccc|}
  \hline
Algorithm  & BBOB & CEC2013 & CEC2014 & CEC2015 \\ 
  \hline
CMA &  0.033 & 0.261 & 0.234 & 0.228  \\ 
PSO & 0.055 & 0.173 & 0.223 & 0.208 \\ 
DE & 0.033 & 0.231 & 0.201 & 0.279 \\ 
 \hline
\end{tabular}
\end{table}

Here are the outcomes derived from assessing performance predictive models for three algorithms across individual benchmark suites:

\noindent\textbf{BBOB} -- 
The models trained on BBOB consistently yield larger errors (compared to the training errors) across all benchmark suites for PSO and CMA-ES, aligning with the anticipated outcomes based on the feature space observations. However, for DE, the errors display variability across the benchmark suites, a trait that might be influenced by the specific behavior of the DE algorithm.
\textbf{CEC2013} -- 
Training models on CEC2013 result in comparable testing errors across all benchmark suites and algorithms. However, in line with the statistical pattern observed in the feature space indicating smaller errors on CEC2014, it is not clearly visible.
\textbf{CEC2014} -- 
The models trained using CEC2014 exhibit anticipated larger errors when evaluated on BBOB. Regarding evaluations on CEC2015, all algorithms showcase smaller errors, in line with the statistical patterns observed in the feature space. Specifically, when evaluating on CEC2013, the PSO and DE algorithms reflect the statistical pattern observed in the feature space, whereas CMA exhibits errors more akin to those obtained on BBOB, differing from the feature space pattern.
\textbf{CEC2015} -- 
Across all three algorithms, we observe a consistency between the outcomes depicted in the landscape feature space and the evaluation of predictive models. Lower errors are evident in CEC2013 and CEC2014, whereas they notably rise in BBOB.

To support our findings, instead of using MDAE (aggregated errors across the entire test suite), we have included box plots displaying absolute errors for individual instances from a specific set of an RF model predicting the performance of CMA, DE, and PSO. Subplot titles denote the benchmark suite used for model training, with one box plot illustrating training absolute errors and the others depicting corresponding test absolute errors (Fig~\ref{fig:pred_errors_boxplots}).

\begin{figure*}
\centering
\begin{subfigure}[b]{0.32\textwidth}
   \includegraphics[width=\linewidth]{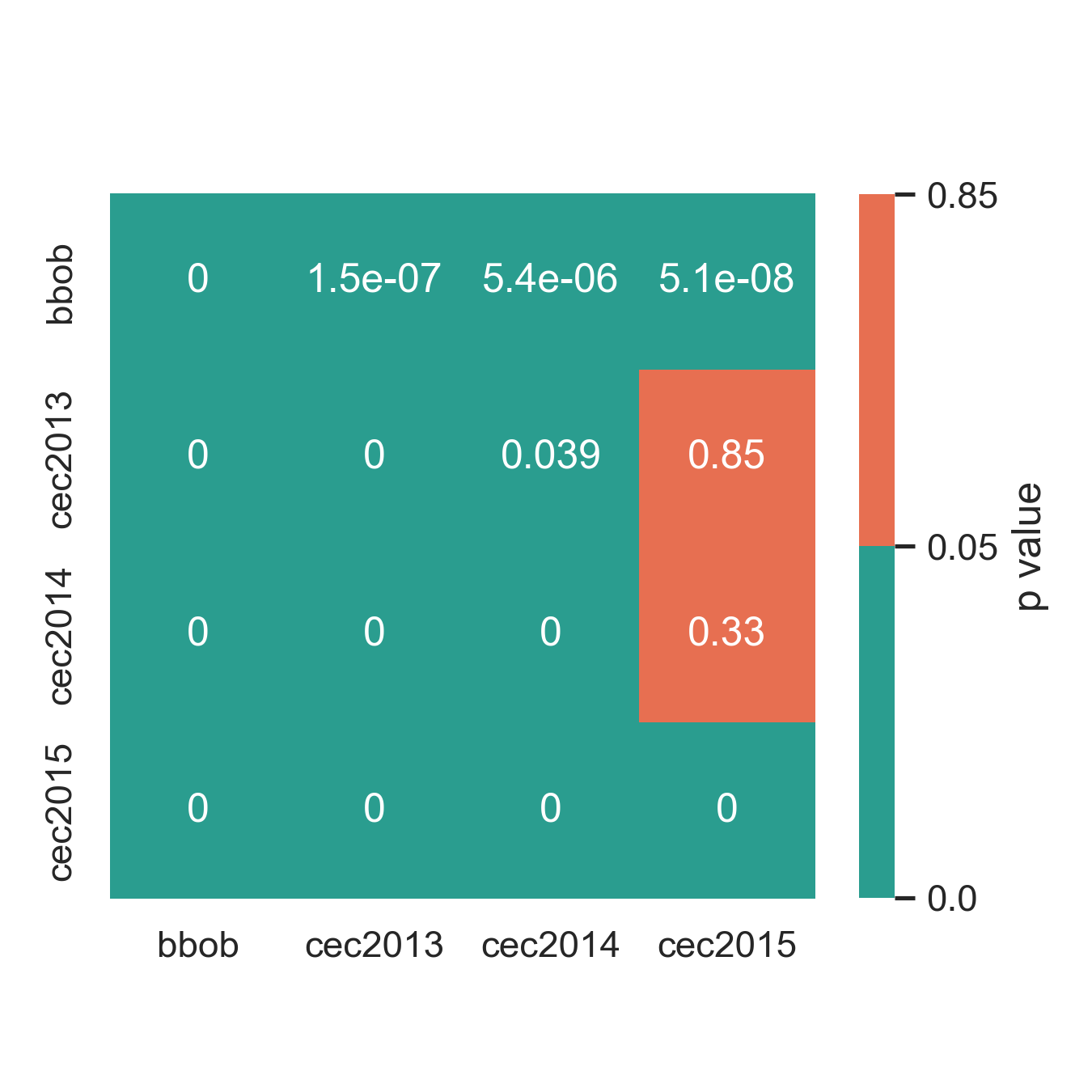}
\hspace{-10mm}
   \caption{CMA}
   \label{fig:CMA} 
\end{subfigure}
\begin{subfigure}[b]{0.32\textwidth}
   \includegraphics[width=\linewidth]{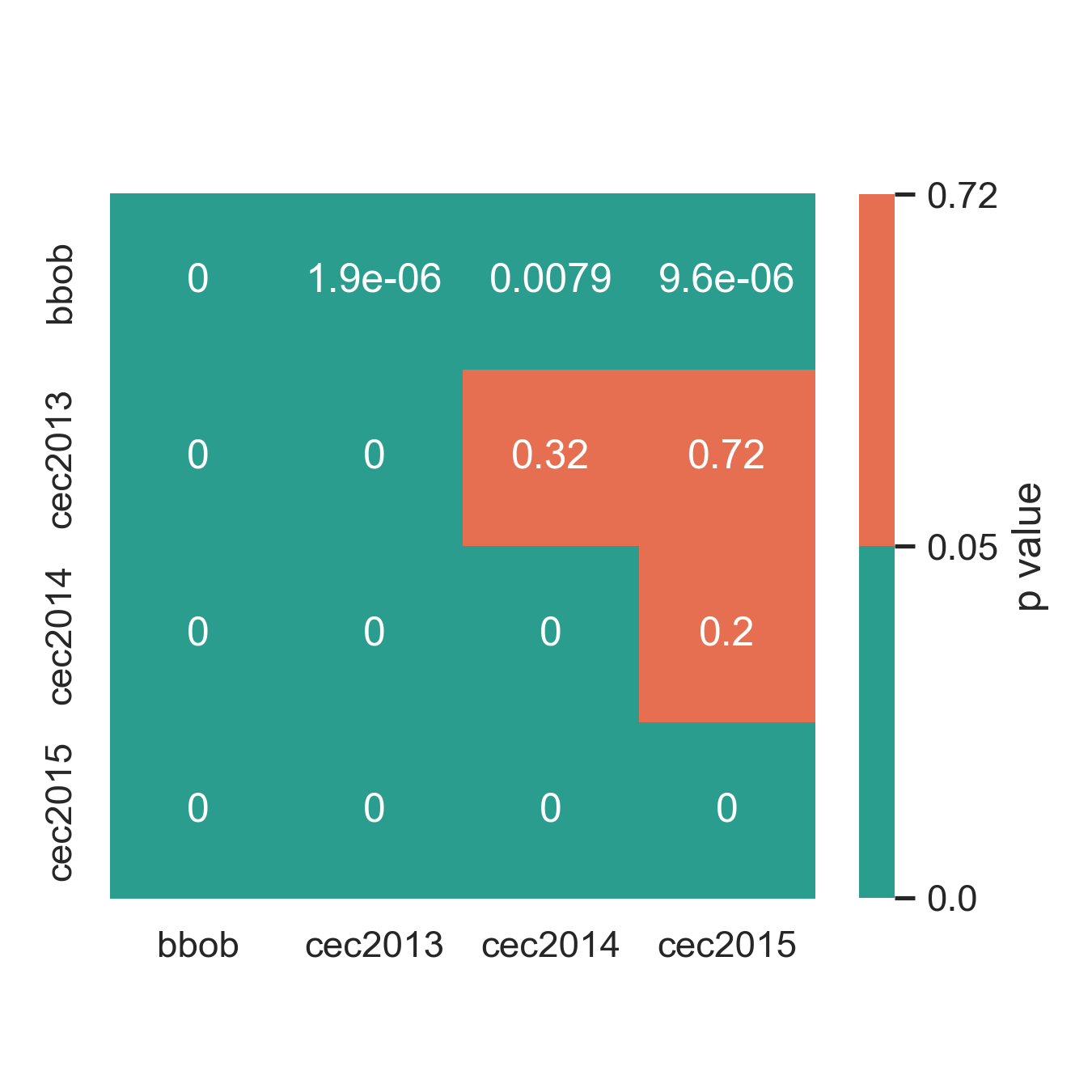}
    \hspace{-5mm}
   \caption{DE}
   \label{fig:DE}
\end{subfigure}
\begin{subfigure}[b]{0.32\textwidth}
   \includegraphics[width=\linewidth]{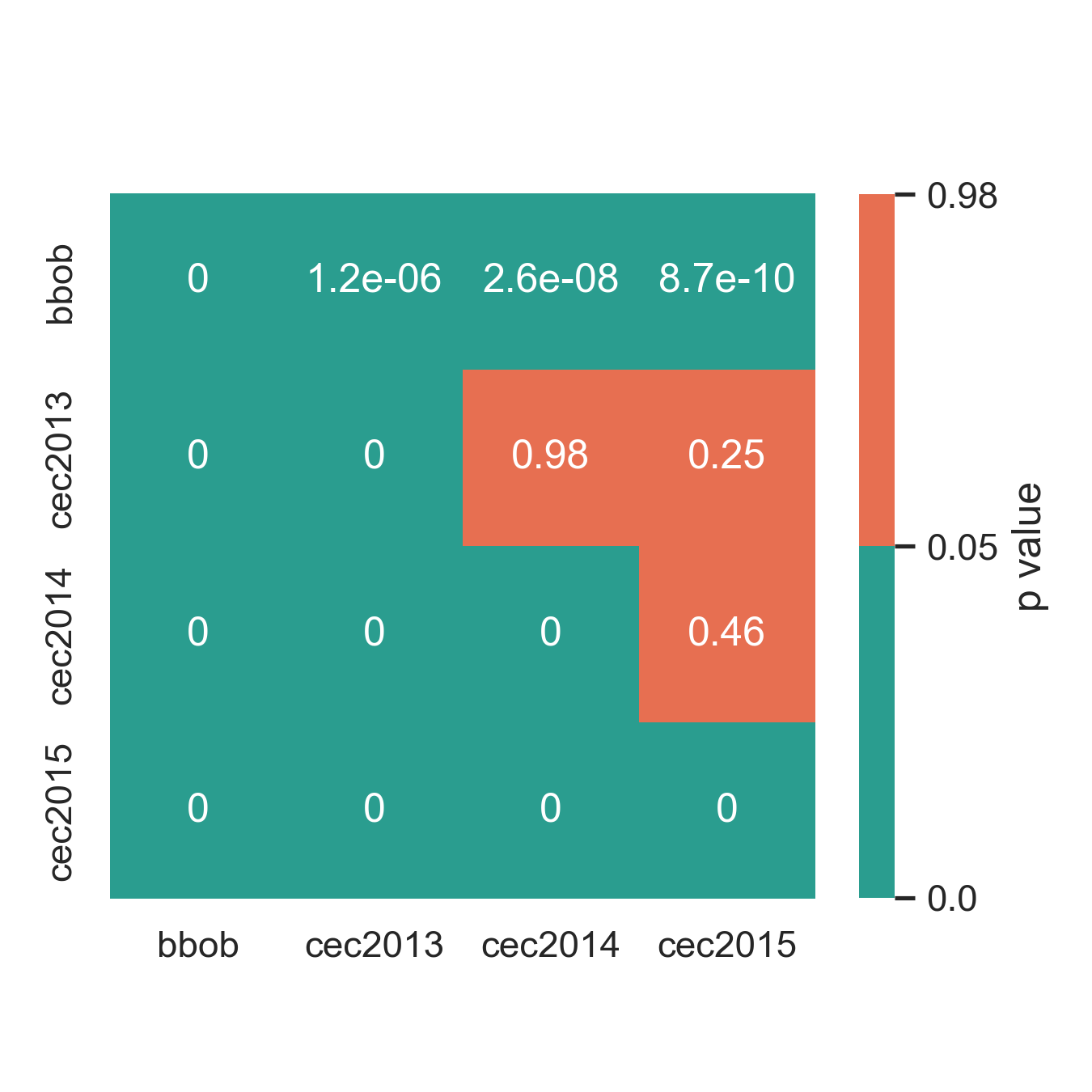}
   \hspace{-5mm}
   \caption{PSO}
   \label{fig:PSO}
\end{subfigure}
\hspace{-5mm}
\caption{
Heatmap that visualizes the p-values obtained by comparing an algorithm's performance distributions among pairs of benchmark suites (a) CMA, b) DE c) PSO). 
Rows and columns depict benchmark suites in paired comparisons. Upper triangle heatmaps show symmetry. A two-sample Kolmogorov-Smirnov test (p-value < 0.05) indicates significant differences in algorithm performance between benchmark suites.}
\label{fig:pvalues}
\vspace{-5mm}
\end{figure*}

The statistical patterns observed in the feature space for BBOB, CEC2014, and CEC2015, generally align with the evaluation of algorithm prediction models. However, this expectation does not hold as strongly for the CEC2013 benchmark suite. 
Additionally, in examining this result, we assess the performance distribution of individual algorithms within each benchmark suite. This involves conducting separate pairwise comparisons of performance distributions across diverse benchmark suites for each algorithm. 
Fig.~\ref{fig:pvalues} displays the p-values resulting from comparing an algorithm's performance distributions by using the two-sample Kolomogorov-Smornov test across pairs of benchmark suites. Each row and column represent benchmark suites used in the pairwise comparison. The heatmaps are in the upper triangle due to the symmetric nature of the comparisons.

If we revisit the CEC2014 case study, the statistical observation in the feature landscape space compared to CEC2013 suggests that these models exhibit generalization ability. However, this is true for the PSO and DE, while it is not in the case for CMA. Upon analyzing the performance distribution comparison, we observe that there is no statistically significant difference between the performance distributions of DE across CEC2014 and CEC2013, the same is true for PSO. However, examining CMA reveals a statistical significance in its performance distributions between CEC2014 and CEC2013. This outcome indicates that despite the statistical similarity found in the feature landscape space of these benchmark suites, such similarity is not evident in the performance space of the CMA algorithm. This outcome suggests that the chosen ELA feature portfolio might lack the capability to detect varied CMA behavior, which is not the case for DE and PSO. 
In the future, we aim to identify specific meta-features tailored to individual algorithms or their respective families. This could involve conducting feature selection on the ELA features or creating and assessing alternative landscape features~\cite{petelin2023tinytla,prager2022automated}. 

\subsection{Second experiment} 
Table~\ref{tab:stat_affine} illustrates the p-values acquired through the pairwise comparisons between two selected benchmark suites, where one suite is utilized for training and the other for testing the model. The outcomes revealed no statistically significant differences among the pairs, as anticipated. This aligns with our expectations because all chosen benchmark suites were sampled using the same technique -- SELECTOR. Based on the results, we anticipate that models trained on one selected benchmark suite will demonstrate good generalization ability when tested on another selected benchmark suite. 
\begin{table}[ht]
\centering
\scriptsize 
{
\caption{Comparative statistical analysis of high-dimensional feature-space distributions between paired benchmark suites artificially sampled from AFFINE problems (presenting p-values).}
\label{tab:stat_affine}
\begin{tabular}{|r|ccccc|}
  \hline
 & BS1 & BS2 & BS3 & BS4 & BS5 \\ 
  \hline
BS1 & / & 0.86 & 0.97 & 0.56 & 0.92 \\ 
  BS2 & 0.90 & / & 0.47 & 0.51 & 1.00 \\ 
  BS3 & 0.99 & 0.88 & / & 0.94 & 0.95 \\ 
  BS4 & 0.98 & 0.89 & 0.99 & / & 0.97 \\ 
  BS5 & 0.93 & 1.00 & 1.00 & 0.98 & / \\ 
   \hline
\end{tabular}
}
\end{table}

\begin{figure*}
\centering
\begin{subfigure}[b]{0.32\textwidth}
   \includegraphics[width=\linewidth]{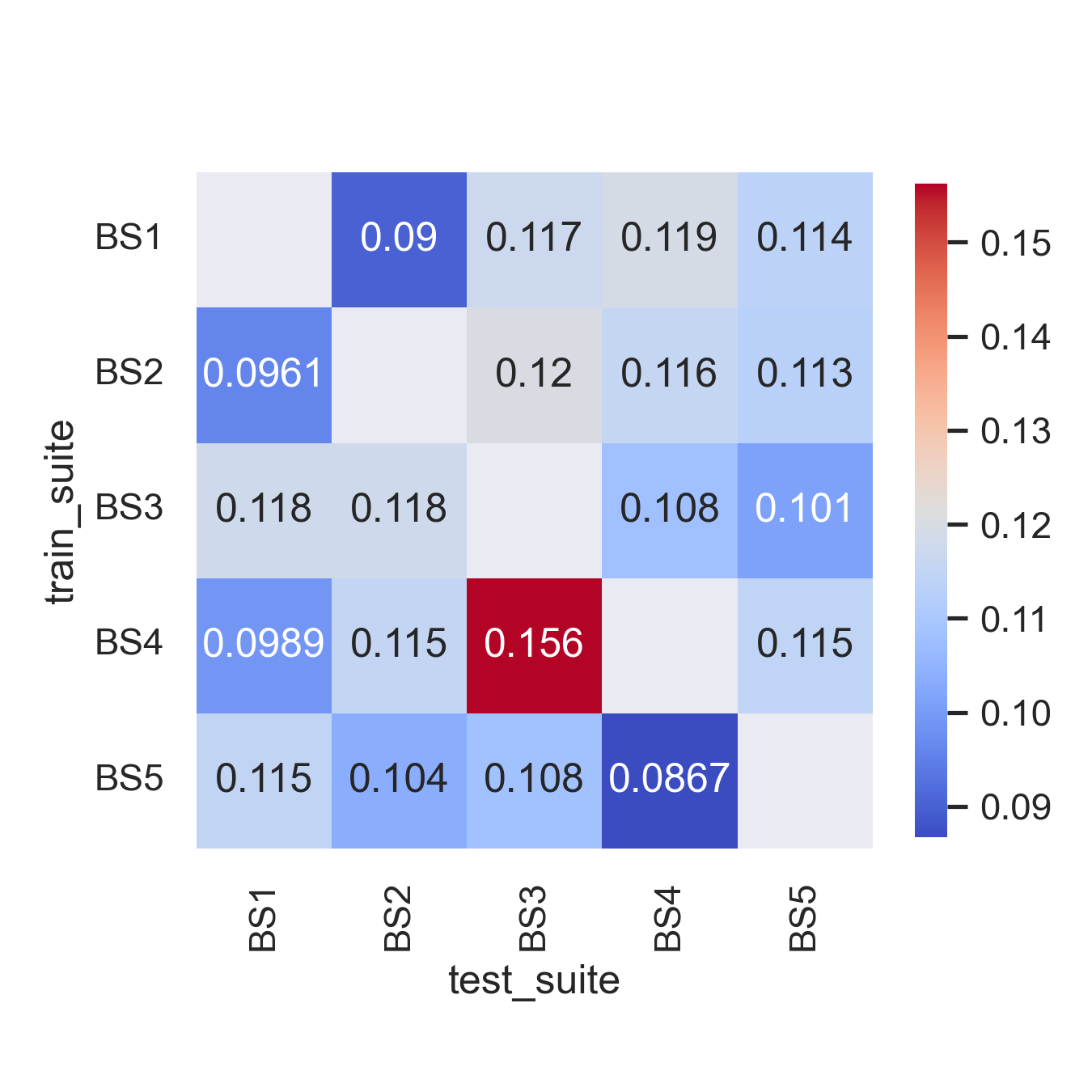}
   \hspace{-5mm}
   \caption{CMA}
   \label{fig:CMA} 
\end{subfigure}
\begin{subfigure}[b]{0.32\textwidth}
   \includegraphics[width=\linewidth]{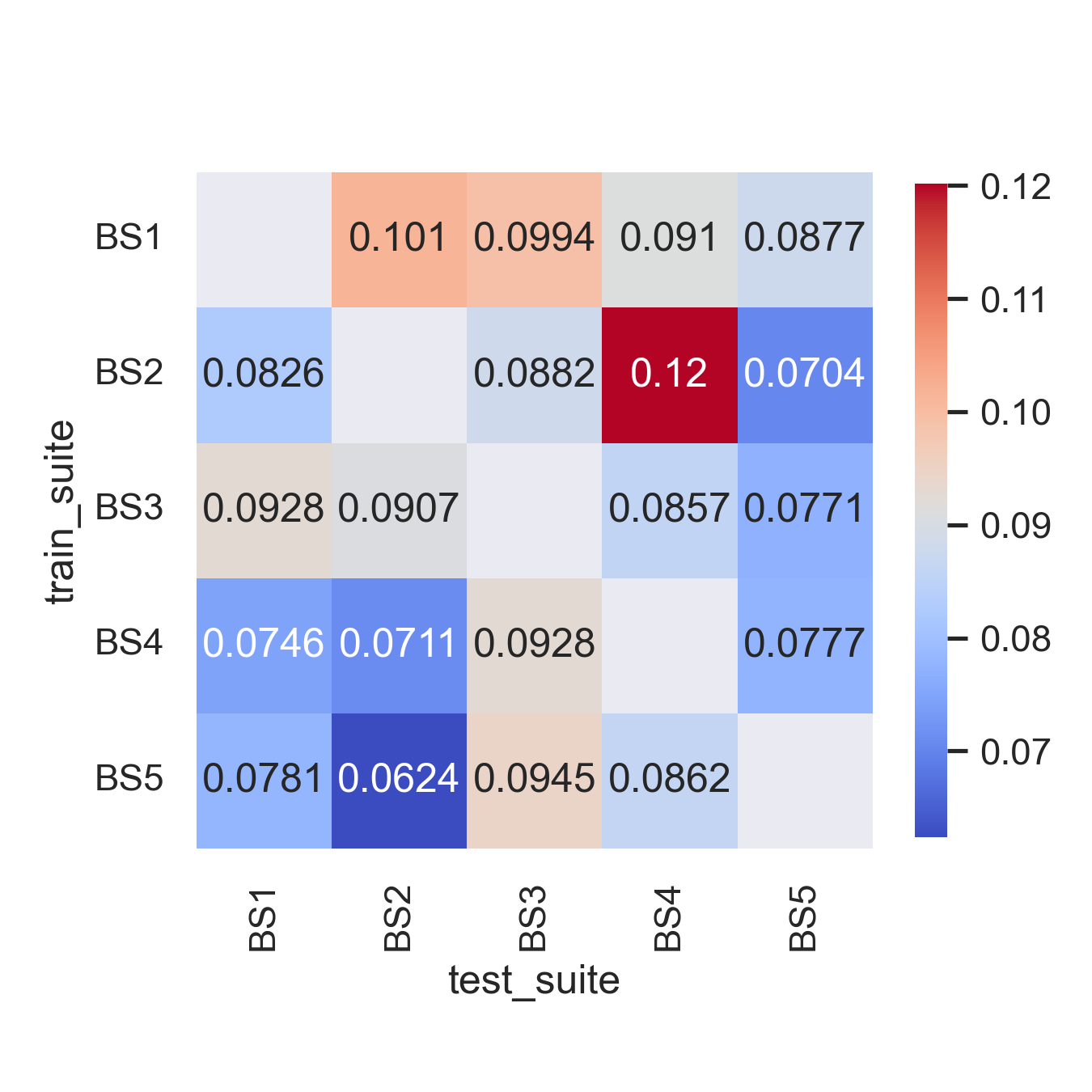}
    \hspace{-5mm}
   \caption{DE}
   \label{fig:DE}
\end{subfigure}
\begin{subfigure}[b]{0.32\textwidth}
   \includegraphics[width=\linewidth]{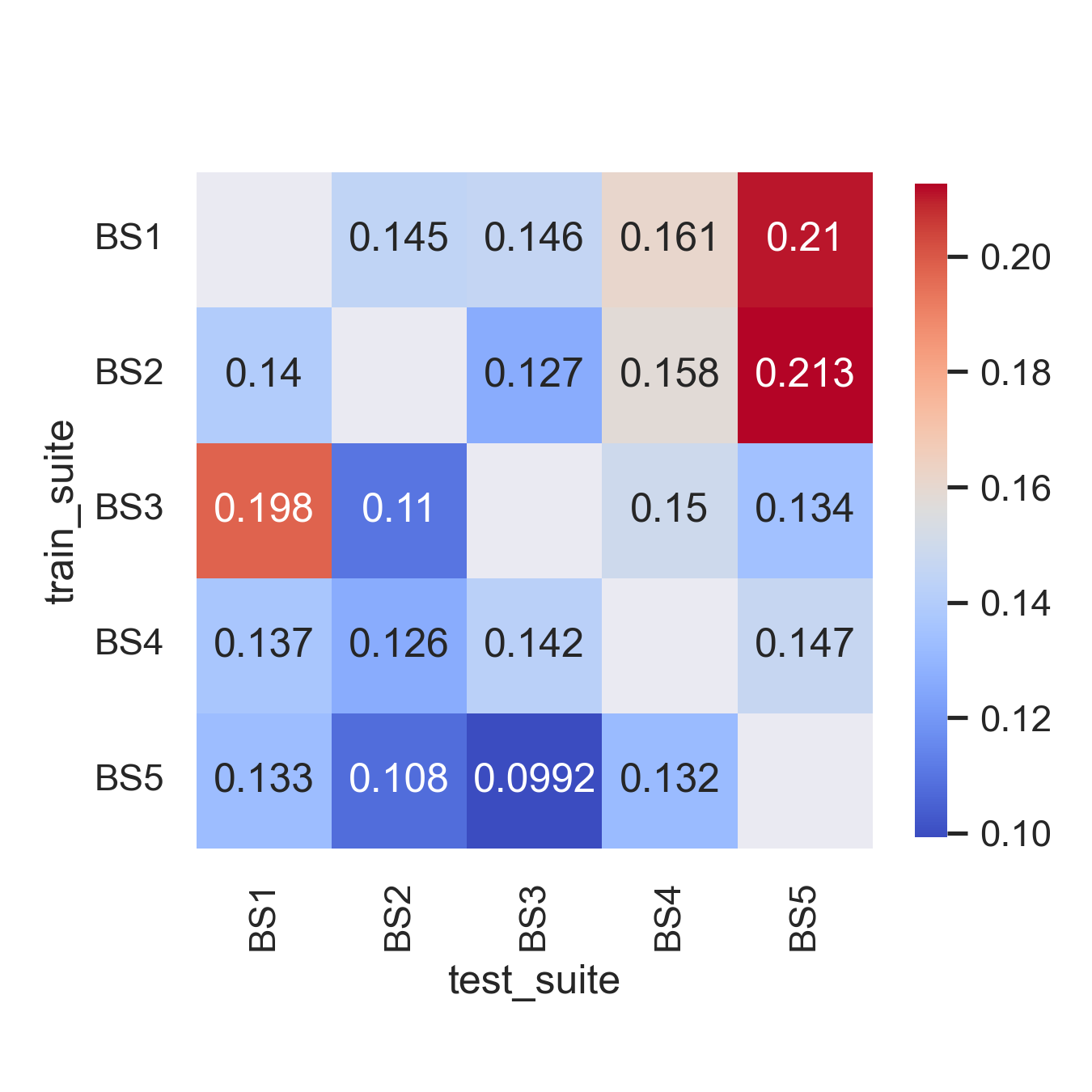}
 \hspace{-5mm}
   \caption{PSO}
   \label{fig:PSO}
\end{subfigure}
\hspace{-5mm}
\caption{Heatmap showing the MDAE of an RF model when predicting the performance of a) CMA, b) DE, and c) PSO on the benchmark suites sampled from the affine problems. Rows indicate the training benchmark suite and columns indicate the benchmark suite of the model was evaluated on.}
\vspace{-5mm}
\label{fig:performance_experiment2}
\end{figure*}

Fig~\ref{fig:performance_experiment2} presents the MDAE of an RF model in predicting the performance of CMA, DE, and PSO across benchmark suites selected from the affine problem instances. Rows represent the training benchmark suite, while columns indicate the evaluated benchmark suite for the model. Based on the findings provided, we can infer that consistent smaller errors are observed among the tested pairs of benchmark suites, mirroring the outcomes showcased in the feature landscape space (i.e.,
the high-dimensional distributions within the feature landscape space of the benchmark suites show no statistical significance.). Moreover, to illustrate that the testing errors are in similar ranges with the training errors, Table~\ref{tab:train_errors_exp_2} displays the training errors of the RF model within each artificially selected benchmark suite from the affine problems for the algorithms Diagonal CMA, PSO, and DE.

\begin{table}[ht]
\centering
\scriptsize 
{
\caption{
The errors during training of the RF model within every benchmark suite artificially sampled from the affine problems for the algorithms Diagonal CMA, PSO, and DE.}
\label{tab:train_errors_exp_2}
\begin{tabular}{|r|ccccc|}
  \hline
Algorithm  & SB1 & SB2 & SB3 & SB4 & SB5 \\ 
  \hline
CMA & 0.0619 & 0.051 & 0.0890 & 0.039 & 0.046 \\ 
PSO & 0.098 & 0.076 & 0.081 & 0.114 & 0.095 \\ 
DE & 0.045 & 0.022 & 0.046 & 0.044 & 0.043 \\ 
 \hline
\end{tabular}
}
\end{table}

\subsection{Discussion}
Since the data used in this study has been taken from another study, we provide a discussion about the results obtained here and the previous results reported. The main difference is the measures used to estimate the generalization ability. In the other study, empirical measures have been introduced. All problem instances from various benchmark suites are aggregated, represented by the same meta-features, and subsequently clustered. To evaluate the similarity among the benchmark suites, a coverage matrix is computed. This matrix quantifies the percentage of problem instances from each suite represented within each cluster. Next, it establishes the benchmark suite meta-representation by utilizing the distribution percentages of instances from each benchmark suite across all clusters. This facilitates the comparison between the meta-representations of two benchmark suites by using similarity measures (e.g., cosine similarity), where one is utilized to train the performance predictive model and the other for testing purposes. Here, instead of using empirical measures, statistical measures are used to estimate the generalization ability. This means that a high-dimensional statistical test is utilized to compare the raw benchmark suite data -- all problem instances represented by their meta-features, without sacrificing information by converting it into a lower-dimensional space, as was done in the empirical case. 

When empirical measures are employed~\cite{10.1145/3583133.3590617}, it becomes evident that all CEC benchmark suites exhibit substantial similarity, displaying minimal differences in the feature space. However, employing innovative statistical measures enables the identification of finer disparities among them, enhancing the accuracy of our expectations regarding model performance errors. Looking ahead, as these measures possess distinct natures and are not directly comparable, employing ensemble techniques can combine their perspectives, leveraging their differing views of the same data.

\section{Conclusions}
\label{sec:conclusion}

This study examined how well a performance predictive model can adapt to new scenarios. We used statistical tests to compare suite coverage distributions in their original high-dimensional feature landscape. By training a model on one benchmark suite and testing it on another, we found that statistical similarities in feature landscape patterns can indicate the model's generalizability. When the distributions between training and testing suites show no significant difference, the model effectively generalizes, maintaining a similar error range. 

In our future work, we plan to conduct a comprehensive experiment using a wider range of algorithms, including Nevergrad~\cite{nevergrad}, to see if the insights gained from the feature landscape analysis extend to broader algorithmic families. Our study demonstrated that performance prediction models built on ELA features effectively generalize across the three tested algorithms. Next, we will explore additional feature landscape meta-features, such as topological features~\cite{petelin2023tinytla} and those derived from deep neural network architectures~\cite{prager2022automated}, comparing them with ELA features to enhance predictive accuracy. Finally, we aim to evaluate these measures in an active learning setting, using them to determine if a model is suitable for new instances or if further training and fine-tuning are necessary.

\bibliographystyle{IEEEtran}
\bibliography{references}

\end{document}